\documentclass[11pt]{article}

\usepackage[margin=1in]{geometry}
\usepackage{amsmath,amssymb}
\usepackage{graphicx}

\graphicspath{{figures/}{./}}

\newcommand{\figfile}[2][width=\textwidth]{%
  \IfFileExists{figures/#2.pdf}{\includegraphics[#1]{figures/#2}}{%
    \IfFileExists{figures/#2.png}{\includegraphics[#1]{figures/#2}}{%
      \IfFileExists{#2.pdf}{\includegraphics[#1]{#2}}{%
        \IfFileExists{#2.png}{\includegraphics[#1]{#2}}{%
          \GenericWarning{}{FIGURE NOT FOUND: tried figures/#2.pdf,
            figures/#2.png, #2.pdf, #2.png}%
          \fbox{\begin{minipage}[c][0.30\textwidth][c]{0.88\textwidth}%
            \centering\ttfamily\detokenize{#2}\par\vspace{4pt}%
            \normalfont\itshape file not found. Looked for
            \texttt{figures/\detokenize{#2}.pdf}, \texttt{.png}, and the same
            two beside \texttt{manuscript.tex}.%
          \end{minipage}}}}}}}
\usepackage{booktabs}
\usepackage{multirow}
\usepackage{xcolor}
\usepackage{tikz}
\usetikzlibrary{positioning,arrows.meta,fit,backgrounds}
\usepackage[hidelinks]{hyperref}
\usepackage{siunitx}
\usepackage{placeins}   
\usepackage{caption}
\usepackage{subcaption}

\newcommand{\headlinetumor}{0.577}
\newcommand{\headlinetumorCI}{[0.518, 0.633]}
\newcommand{\headlineLiver}{0.947}
\newcommand{\headlineLiverCI}{[0.941, 0.952]}
\newcommand{\headlinetumorBearing}{0.565}

\newcommand{\headlineNVol}{131}
\newcommand{\headlineNBearing}{118}

\newcommand{\splittumor}{0.576 $\pm$ 0.044}

\title{\textbf{Parameter-Efficient 3D Segmentation of Liver and Liver tumors:
Depthwise factorization Scales Better Than Dense Convolution
with Spatial Dimensionality}}

\author{%
  Adham M. Alkhadrawi\textsuperscript{1,*} \quad
  Mohammed A.B. Mahmoud\textsuperscript{2,*}\\[8pt]
  \footnotesize \textsuperscript{1}Department of Molecular Biosciences and
  Bioengineering, University of Hawaii at Manoa, Honolulu, HI, USA\\
  \footnotesize \textsuperscript{2}Faculty of Computer Studies,
  Arab Open University, Riyadh 11681, Kingdom of Saudi Arabia\\[4pt]
  \footnotesize \textsuperscript{*}These authors contributed equally. \\
  \footnotesize Corresponding author: Mohammed A.B. Mahmoud,
  \texttt{[m.mahmoud@arabou.edu.sa]}
}
\date{}

\begin{document}
\maketitle

\begin{abstract}
Three-dimensional dense convolutional networks are the strongest performers on
volumetric medical image segmentation, but their parameter counts scale poorly:
moving a dense $k \times k$ convolution to $k \times k \times k$ multiplies its
weights by $k$. We observe that depthwise separable factorization does not share
this penalty. Because the cubic kernel term applies only to the depthwise stage
while the pointwise projection, which dominates the parameter count, is
unchanged, the same architecture grows by \SI{5}{\percent} from 2D to 3D where a
dense convolutional U-Net grows by \SI{200}{\percent}. We exploit this asymmetry
to build a 3D U-Net with \num{536990} parameters, \num{24}$\times$ fewer than an identical dense 3D U-Net.

On MSD Task03\_Liver (the Medical Segmentation Decathlon liver task, derived
from LiTS), evaluated per case under five-fold cross-validation over all
\headlineNVol{} public volumes, the same protocol nnU-Net reports for this
task, the model reaches a tumor Dice of \headlinetumor{} (95\%~CI
\headlinetumorCI{}) and a liver Dice of \headlineLiver{} (95\%~CI
\headlineLiverCI{}). Its liver Dice exceeds every nnU-Net configuration on this
task. Its tumor Dice exceeds their low-resolution configuration (0.4701) by
$0.107$ and their 2D configuration (0.5394), and falls $0.040$ short of their
single-stage 3D U-Net (0.6174), a gap within our confidence interval, at
approximately one twenty-fourth of the parameters and roughly half the in-plane
resolution. Trained under identical
conditions on a common held-out split, it exceeds a dense 3D U-Net on liver by
$+0.031$ Dice (paired $p=0.006$) and on tumor by $+0.041$
(\SI{95}{\percent}~CI $[+0.005, +0.081]$, paired $p=0.056$), suggesting the
factorization also acts as a regularizer in the small-data regime
characteristic of medical imaging. We further show, on both LiTS and a 2D
endoscopy benchmark, that a large fraction of the network's learnable spatial
filters can be replaced by fixed shifts at no cost in accuracy, but that
replacing all of them is measurably worse, the placement of spatial capacity
matters more than its total amount.
\end{abstract}

\FloatBarrier
\section{Introduction}

Automated segmentation of the liver and of hepatic lesions from CT helps to support 
surgical planning, treatment response assessment and tumor burden
quantification. Since the LiTS challenge \cite{bilic2023liver}, the liver/tumor segmenation
task has become a standard benchmark. The strongest reported results come
from three-dimensional convolutional networks, which learns through 3D
context that 2D models cannot.

That performance comes with a cost in model size. A dense $3\times3$ convolution
between $C_\text{in}$ and $C_\text{out}$ channels holds
$9\,C_\text{in}C_\text{out}$ weights; its 3D counterpart holds
$27\,C_\text{in}C_\text{out}$. Every convolutional layer triples. A standard 3D
U-Net at the width used here holds \num{12946851} parameters, and configurations
in the literature reach tens of millions.

Parameter count is not the only thing that determines model efficiency. It does,
however, determines model file size, and therefore the cost of distributing,
and deploying a trained network in hospital settings.

Research on lightweight segmentation networks has focused almost entirely on 2D
tasks \cite{valanarasu2022unext,tang2024cmunext,dinh20231m,ding2025lsu}. We argue that 3D is where
the factorization carries the greatest advantage, for a structural reason. In a depthwise
separable block, the spatial and channel-mixing operations are separated: a
depthwise convolution applies one $k\times k$ (or $k\times k\times k$) filter
per channel, and a $1\times1$ pointwise convolution mixes channels. The
parameter cost is
\begin{equation}
  \underbrace{C_\text{in} \cdot k^d}_{\text{depthwise}}
  \;+\;
  \underbrace{C_\text{in} \cdot C_\text{out}}_{\text{pointwise}},
  \label{eq:dwsep}
\end{equation}
for spatial dimensionality $d$. Only the first term depends on $d$, and it is
the smaller term at realistic channel widths. Moving from
$d{=}2$ to $d{=}3$ therefore never touches the dominant cost. In contrast, A dense
convolution, costs $C_\text{in}C_\text{out}k^d$ throughout, and
scales by the full factor of $k$.

The attempts to reduce this parameter cost has mainly been architectural. Separable formulations reduce the burden by decomposing the 3D kernel: S3D-UNet splits each convolution into three parallel branches along orthogonal anatomical views (sagittal, coronal, and axial) \cite{chen2018s3d}, while lightweight attention variants use parallel convolutions in combination with attention mechanisms \cite{alwadee2025latup}. A different approach avoids volumetric convolution completely, exploiting the parameter efficiency of a 2D network applied across multiple planes and recombined at test time \cite{perslev2019one}.

\paragraph{Contributions.}
\begin{enumerate}
  \item We identify and quantify a scaling asymmetry: depthwise separable
    factorization grows by \SI{5}{\percent} from 2D to 3D at fixed architecture
    and width, where dense convolution grows by \SI{200}{\percent}
    (Section~\ref{sec:scaling}). The advantage of the factorization therefore
    increases with spatial dimensionality.
  \item We build a 3D U-Net of \num{536990} parameters on this principle and
    evaluate it on LiTS with a volume-level split and per-case Dice, reaching
    \headlinetumor{} tumor Dice, comparable to published 3D configurations at
    roughly one twenty-fifth of the parameters, and better than a dense 3D U-Net
    trained identically on the same data.
  \item We characterize, through an $L_0$-gated variant and a family of
    partially frozen models, \emph{where} in a U-Net learnable spatial filtering
    is required. Most of it can be replaced by parameter-free fixed shifts
    without loss, however, replacing all of it cannot. This is true in both a 2D and a 3D
    benchmark (Section~\ref{sec:allocation}).
  \item We report measured, rather than assumed, consequences of parameter
    reduction: unchanged training memory and a \num{1.2}$\times$ CPU inference
    speed-up, alongside the \num{24}$\times$ reduction in model size
    (Section~\ref{sec:cost}).
\end{enumerate}

\FloatBarrier
\section{Related Work}

\subsection{Parameter-efficient segmentation networks}
A substantial amount of research work focused on the size of U-Net-based segmentation models.
UNeXt \cite{valanarasu2022unext} replaces deep convolutional stages with tokenized
MLP blocks. CMUNeXt \cite{tang2024cmunext} and U-Lite \cite{dinh20231m}
use large-kernel depthwise convolutions with narrow channel schedules to reach
under one million parameters. LSU-Net \cite{ding2025lsu} combines a
fixed spatial-shift operator with depthwise separable convolutions and
multi-scale deep supervision. These models are validated on 2D benchmarks; to
our knowledge the scaling behavior that motivates the present work has not been
stated explicitly, and lightweight designs have not been carried to 3D liver
tumor segmentation.

\subsection{Depthwise separable and shift-based operators}
Depthwise separable convolution was introduced in Xception
\cite{chollet2017xception} and popularized by MobileNet \cite{howard2017mobilenets}.
Shift operations \cite{wu2018shift} replace learned spatial filtering
with parameter-free translation of channel groups, and have been refined to
learnable offsets \cite{jeon2018constructing} and sparse variants
\cite{chen2019all}. Spatial-shift MLPs \cite{yu2021s}
supply the shift formulation adopted by LSU-Net. Our fixed-shift baseline is a
depthwise filter constrained to a one-hot kernel, which makes shift and
depthwise convolution two points on a single continuum rather than distinct
operators (Section~\ref{sec:allocation}).

MobileNetV2 established the inverted-residual block with linear bottlenecks \cite{sandler2018mobilenetv2}, and ShuffleNet showed that grouped pointwise convolution with channel shuffling recovers cross-channel mixing at a fraction of the cost \cite{zhang2018shufflenet}. Transfer to volumetric segmentation has been partial and inconsistent. S3D-UNet separates spatially, into orthogonal 2D views, rather than separating spatial filtering from channel mixing \cite{chen2018s3d}. Both decompositions reduce parameters, but they trade different things. we argue that only channel-spatial separation isolates a depthwise stage which is small enough to be replaced with a fixed shift costing almost nothing.

\subsection{Liver and liver tumor segmentation}
The LiTS challenge \cite{bilic2023liver} established the benchmark
used in this study. Cascaded approaches, which localize the liver before segmenting
lesions within it, were prominent among early studies
\cite{christ2016automatic,li2018h}. nnU-Net \cite{isensee2021nnu}
reported 2D, 3D full-resolution, 3D low-resolution and cascaded configurations on
LiTS, and remains the reference point for methodological comparison; we compare
against its published figures throughout.

Reported LiTS figures are not directly comparable across papers.
Challenge submissions are scored on 70 withheld test cases through the
evaluation server; most subsequent work reports cross-validation over the 131
public training cases; and some report a single held-out split. We mention the
protocol for every number we cite (Table~\ref{tab:benchmark}).

\FloatBarrier
\section{Methods}

\subsection{The scaling asymmetry}
\label{sec:scaling}

Table~\ref{tab:scaling} instantiates Equation~\ref{eq:dwsep} for the U-Net
topology used in this work. Parameter counts are obtained by
construction rather than estimation, at a matched configuration (single input
channel, three output classes, base width 64) so that the 2D and 3D columns
differ only in the dimensionality of the convolution.

\begin{table}[htbp]
\centering
\caption{Parameter count of an identical U-Net topology under 2D and 3D
convolution. The depthwise separable variant pays the cubic kernel term only on
the depthwise stage; the pointwise projection, which dominates, is unchanged.}
\label{tab:scaling}
\begin{tabular}{lrrr}
\toprule
Convolution & 2D & 3D & Growth \\
\midrule
Dense $k^d$                  & \num{17266371} & \num{51772227} & $3.00\times$ \\
Depthwise separable          & \num{1978252}  & \num{2073886}  & $1.05\times$ \\
\bottomrule
\end{tabular}
\end{table}

The asymmetry follows directly from Equation~\ref{eq:dwsep}: the depthwise term
$C_\text{in}k^d$ triples from $d{=}2$ to $d{=}3$, but at the widths used here it
accounts for a small fraction of the block. At $C_\text{in}{=}C_\text{out}{=}512$
and $k{=}3$, the depthwise term is \num{13824} weights against
\num{262144} for the pointwise projection.

Since activation memory rather than parameter count binds in 3D, we adopt a
base width of 32 channels rather than the 64 conventional in 2D
(Section~\ref{sec:cost}).

\begin{figure}[htbp]
\centering
\figfile[width=\textwidth]{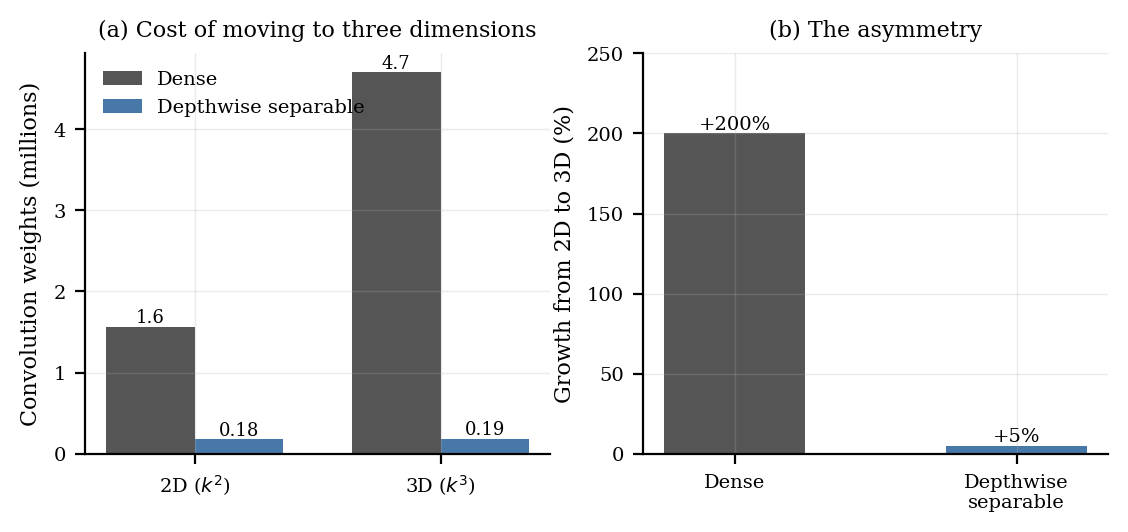}
\caption{The scaling asymmetry, computed from
Equation~\ref{eq:dwsep} over the four encoder stages of the architecture in
Figure~\ref{fig:arch}. (a) Moving a dense U-Net from $k\times k$ to
$k\times k\times k$ kernels multiplies its convolution weights by $k$. The
depthwise separable form does not pay this, because the cubic term applies only
to the depthwise stage while the pointwise projection, which dominates the
count, is unchanged. (b) The same data as growth rates: \SI{200}{\percent}
against \SI{5}{\percent}.}
\label{fig:scaling}
\end{figure}

\subsection{Architecture}
\label{sec:arch}

The backbone is a five-level U-Net \cite{ronneberger2015u,cciccek20163d} with trilinear upsampling, instance
normalization and ReLU. Instance normalization \cite{ulyanov2016instance} is used in place of batch
normalization because 3D patch training operates at batch size 2, where batch
statistics are unreliable \cite{isensee2021nnu}. Each level applies two
convolutional blocks; we compare only the block differs between the variants.

Figure~\ref{fig:arch} shows the network and the three block variants.

\begin{figure}[htbp]
\centering
\resizebox{0.92\textwidth}{!}{

\begin{tikzpicture}[
  font=\small,
  lvl/.style   = {draw, rounded corners=2pt, minimum width=26mm,
                  minimum height=7mm, align=center, fill=blue!8},
  bott/.style  = {lvl, fill=blue!18},
  op/.style    = {draw, rounded corners=2pt, minimum width=30mm,
                  minimum height=7mm, align=center},
  spatial/.style = {op, fill=orange!20},
  point/.style   = {op, fill=teal!18},
  norm/.style    = {op, fill=gray!12},
  frozen/.style  = {op, fill=gray!30, draw=black, dashed},
  arr/.style   = {-{Latex[length=2mm]}, thick},
  skip/.style  = {-{Latex[length=2mm]}, thick, dashed, gray!70},
  lab/.style   = {font=\scriptsize\itshape, text=black!60}
]

\node[font=\bfseries] at (-1.2, 0.9) {A};

\node[lvl]  (e1) at (0,  0)    {$32$};
\node[lvl]  (e2) at (0, -1.3)  {$64$};
\node[lvl]  (e3) at (0, -2.6)  {$128$};
\node[lvl]  (e4) at (0, -3.9)  {$256$};
\node[bott, minimum width=34mm] (b) at (3, -5.4) {$256$};
\node[lab] at (3, -6.0) {bottleneck};

\node[lvl]  (d4) at (6, -3.9)  {$128$};
\node[lvl]  (d3) at (6, -2.6)  {$64$};
\node[lvl]  (d2) at (6, -1.3)  {$32$};
\node[lvl]  (d1) at (6,  0)    {$32$};

\node[op, fill=green!14, minimum width=20mm] (out) at (9, 0) {$1\times1\times1$\\ 3 classes};

\draw[arr] (e1) -- node[lab, right=1pt] {pool} (e2);
\draw[arr] (e2) -- node[lab, right=1pt] {pool} (e3);
\draw[arr] (e3) -- node[lab, right=1pt] {pool} (e4);

\draw[arr] (e4.south) |- (b.west);
\draw[arr] (b.east) -| (d4.south);
\draw[arr] (d4) -- node[lab, right=1pt] {up} (d3);
\draw[arr] (d3) -- node[lab, right=1pt] {up} (d2);
\draw[arr] (d2) -- node[lab, right=1pt] {up} (d1);
\draw[arr] (d1) -- (out);

\draw[skip] (e1) -- (d1);
\draw[skip] (e2) -- (d2);
\draw[skip] (e3) -- (d3);
\draw[skip] (e4) -- (d4);

\node[draw=orange!70, dashed, thick, rounded corners=3pt, inner sep=2.2pt,
      fit=(d4)] (hl) {};
\node[lab, anchor=west, text=orange!60!black] at (7.5, -3.9)
  {\parbox{30mm}{\raggedright \textbf{up1}: first decoder stage,\\
   fixed shifts in variant (c)}};
\draw[orange!70, dashed, thick] (hl.east) -- (7.45, -3.9);

\node[lab] at (3, 0.55) {each level = 2 blocks (Panel B) + norm + ReLU};
\node[lab] at (3, -6.7) {input: $1 \times 128^3$ patch \quad output: $3 \times 128^3$};

\begin{scope}[yshift=-9.2cm]
\node[font=\bfseries] at (-1.2, 1.4) {B};

\node[font=\bfseries\small] at (0.4, 1.0) {(a) Fixed shift};
\node[frozen]  (a1) at (0.4,  0.2) {shift $3^3$, one-hot\\ \scriptsize not trained};
\node[point]   (a2) at (0.4, -1.1) {pointwise $1^3$};
\draw[arr] (a1) -- (a2);
\node[lab, align=center] at (0.4, -2.1)
  {spatial params: \textbf{0}\\ \num{485987} total};

\node[font=\bfseries\small] at (5.0, 1.0) {(b) Learnable depthwise};
\node[spatial] (b1) at (5.0,  0.2) {depthwise $3^3$\\ \scriptsize trained};
\node[point]   (b2) at (5.0, -1.1) {pointwise $1^3$};
\draw[arr] (b1) -- (b2);
\node[lab, align=center] at (5.0, -2.1)
  {spatial params: $C\!\cdot\!27$\\ \num{557726} total};

\node[font=\bfseries\small] at (9.6, 1.0) {(c) Hybrid};
\node[spatial] (c1) at (9.6,  0.2) {depthwise $3^3$\\ \scriptsize trained, most stages};
\node[frozen]  (c1b) at (9.6, -1.1) {\scriptsize fixed shift in the\\ \scriptsize first decoder stage};
\node[point]   (c2) at (9.6, -2.4) {pointwise $1^3$};
\draw[arr] (c1) -- (c1b);
\draw[arr] (c1b) -- (c2);
\node[lab, align=center] at (9.6, -3.4)
  {\num{536990} total};

\node[lab, align=left, anchor=west] at (-1.0, -4.6)
  {\parbox{125mm}{\raggedright
   A one-hot depthwise kernel \emph{is} a shift, so (a) and (b) are the two
   endpoints of one continuum---the number of non-zero taps per filter, from 1
   to $k^3$---rather than distinct operators. The pointwise stage is identical
   in all three and dominates the parameter count, which is why the variants
   sit within \SI{15}{\percent} of each other despite differing in whether
   spatial filtering is learned at all.}};
\end{scope}

\end{tikzpicture}}
\caption{\textbf{(A)} The 3D U-Net backbone. Numbers give the channel width at
each level; the base width of 32 is halved relative to the 64 conventional in
2D because activation memory, not parameter count, binds in 3D. Dashed arrows
are skip connections. \textbf{(B)} The three convolutional block variants
compared in this work. All share the structure spatial mixing followed by
channel mixing and differ only in the spatial stage, so any difference in
accuracy is attributable to that stage alone.}
\label{fig:arch}
\end{figure}

\subsubsection{Three block variants}

All three share the structure \emph{spatial mixing} $\rightarrow$ \emph{channel
mixing}, and differ only in what the spatial stage is and whether it is learned.
This makes them directly comparable at nearly matched
parameter counts, so any difference is attributable to the spatial operator
alone.

\paragraph{(a) Fixed shift.}
The spatial stage is a depthwise convolution whose kernel is constrained to be
one-hot: each channel is translated by one voxel in one of the 26 neighbor
directions of a $3\times3\times3$ neighborhood, with a 27th group left
stationary. The weights are not trained. This is exactly a shift operator, and
it costs zero spatial parameters. Channels are assigned to direction groups in
contiguous blocks.

We note that a common implementation of shift blocks pads the channel count to a
multiple of the number of shift groups using a $1\times1$ projection. That
projection is not required by the operation, the remainder can be assigned to
the stationary group, and it is expensive: in our 2D reproduction it accounted
for \num{2728347} of the \num{4658910} parameters, \SI{59}{\percent} of the
model, for no measurable accuracy benefit (Section~\ref{sec:results-2d}). Therefore, we omit that padding in Our fixed-shift baseline.

\paragraph{(b) Learnable depthwise.}
The spatial stage is an unconstrained depthwise $3\times3\times3$ convolution:
each channel receives its own trainable filter. This is the standard depthwise
separable block \cite{chollet2017xception,howard2017mobilenets}. Relative to (a) it adds
$C_\text{in}k^3$ parameters per layer.

\paragraph{(c) Hybrid.}
Learnable depthwise filters throughout, except in \texttt{up1}, the first
decoder stage, fed directly by the bottleneck, which is constrained to fixed
shifts as in (a).

\subsubsection*{Why \texttt{up1}}
\label{sec:whyup1}

Three independent considerations identify the first decoder stage as the natural
place to remove learnable spatial filtering.

\begin{enumerate}
  \item \textbf{The measured allocation points there.} Under $L_0$ pressure the
    deep decoder falls to fewer than two open taps per filter, close to the
    single tap that constitutes a shift, while the encoder holds above five
    (Table~\ref{tab:taps}). The ordering is reproducible across seeds, with a
    per-layer standard deviation of 0.16, and strengthens consistently as the
    penalty increases. \texttt{up1} is the deepest decoder stage and the one the
    gates release first.

  \item \textbf{\texttt{up1} is the most expensive stage to leave learnable.}
    Depthwise cost scales with channel width, and \texttt{up1} is the widest decoder stage ($512 \rightarrow 256 \rightarrow 128$
    channels). Freezing it alone removes \num{20736} parameters,
    \SI{28.9}{\percent} of every depthwise weight in the network, more than
    twice the next single stage (\texttt{down4}, \num{13824}) and more than the
    four shallowest stages combined. It is therefore the single intervention
    that converts the measured redundancy into the largest saving.

  \item \textbf{It operates where the least spatial detail remains.} For a
    $128^3$ input, \texttt{up1} acts on feature maps at $16^3$ resolution, one
    step above the $8^3$ bottleneck. A $3\times3\times3$ filter there spans a
    large physical extent of the volume, and the feature map carries
    correspondingly little fine structure for a learned spatial filter to
    discriminate. Fine spatial detail is recovered in the shallow decoder
    stages, which we leave learnable.
\end{enumerate}

\noindent Together, these 3 reasons agree, and they motivate our choice of
variant~(c).

We also report how sensitive the result is to this choice. On the 2D benchmark
we swept five freezing configurations spanning \SI{29}{\percent} to
\SI{84}{\percent} of filters frozen, at five seeds each
(Table~\ref{tab:cutpoints}). Every configuration exceeded the fully frozen
model, and no configuration differed significantly from any other (one-way ANOVA
$F=0.61$, $p=0.66$). We therefore conclude that: \emph{some} learnable spatial filtering must be retained, most of it need
not be, and performance is not sharply sensitive to which stages are released.
\texttt{up1} is the choice that maximizes the saving within that finding.

This sensitivity analysis was performed in 2D. We have not repeated it in 3D,
where the deepest feature maps are far smaller, and we do not assume the
insensitivity transfers.

\subsubsection{Where is learnable spatial filtering needed?}
\label{sec:allocation}

A one-hot depthwise filter \emph{is} a shift. Variants (a) and (b) are therefore
not distinct architectures but the two endpoints of a single continuum: the
number of non-zero taps per filter, from 1 (a shift) to $k^d$ (a full depthwise
filter). The question then follows, how many taps each layer requires?

\begin{equation}
\mathcal{L} \;=\; \mathcal{L}_{\text{seg}}
\;+\; \lambda \sum_{j=1}^{J} P(z_j \neq 0),
\label{eq:2}
\end{equation}

Where $\mathcal{L}$ is the total loss, $\mathcal{L}_{\text{seg}}$ is the segmentation loss.

We add a hard-concrete $L_0$ gate \cite{louizos2017learning} to each tap of
each depthwise filter, so that the effective kernel is $w \odot z$ with $z$
stochastic and a penalty $\lambda\sum_j P(z_j \neq 0)$ applied to the expected
number of open taps (Equation~\ref{eq:2}). The penalty is the probability that a
gate is open rather than its expected magnitude, since the $L_0$ norm counts
non-zero entries; the hard-concrete relaxation makes that count differentiable
by placing finite probability mass at exactly zero. A floor of one open tap per
filter, not shown in Equation~\ref{eq:2}, prevents the penalty from deleting
channels outright, which would be channel pruning rather than the shift
endpoint. We swept $\lambda$ to trace the whole continuum within a single
implementation, and the learned tap budget per layer is a direct measurement of
where spatial capacity is used. The resulting allocation
(Section~\ref{sec:results-allocation}) motivates variant (c).

\subsection{Data and preprocessing}

We use the 131 publicly annotated volumes of Task03\_Liver, the liver task of
the Medical Segmentation Decathlon \cite{antonelli2022medical}, which is derived from
the LiTS challenge training set \cite{bilic2023liver}. We refer to it
as MSD Liver throughout, since nnU-Net reports the MSD liver task and the LiTS
challenge as separate entries and our comparison is with the former. Volumes are resampled
to \SI{1.5}{\milli\meter} isotropic spacing, trilinear for images, nearest
neighbor for labels, so that a fixed-size patch covers a constant physical
region across scans whose native spacing varies widely: slice thickness ranges
from \SI{0.70}{\milli\meter} to \SI{5.0}{\milli\meter} across the
\headlineNVol{} volumes, and in-plane spacing from \SI{0.56}{\milli\meter} to
\SI{1.0}{\milli\meter}, with a median of $1.0 \times 0.77 \times
0.77$~\si{\milli\meter}. Intensities are clipped to $[-17, 201]$~HU and normalized
as $(x - 99.40)/39.39$. We applied these four constants as fixed preprocessing parameters, applied identically to
every volume and to every model variant compared in this study; the clipping range
concentrates dynamic range on liver parenchyma and lesions at the cost of
discarding bone and lung detail. They are recorded in the preprocessing
manifest released with the code.

For the ablation study, the 131 volumes are split at the \emph{volume} level into 93 training, 19
validation and 19 test cases. Splitting at the slice level, as is sometimes
done, places adjacent slices of the same patient in both training and test sets
and inflates results; we quantify this effect in
Section~\ref{sec:results-protocol}.

\subsection{Training}

Models are trained on random $128^3$ patches at batch size 2, using AdamW
\cite{loshchilov2017decoupled} (learning rate $10^{-3}$, weight decay
$10^{-5}$) with a cosine schedule \cite{loshchilov2016sgdr} annealing to
$10^{-5}$ over 200 epochs of 250 sampled patches. Mixed precision
(bfloat16) is used throughout.

\paragraph{Foreground oversampling.}
Across the \headlineNVol{} preprocessed volumes tumor occupies
\SI{0.103}{\percent} of all voxels and \SI{4.52}{\percent} of liver voxels
(\num{2.94e9} voxels in total). Uniformly sampled
patches would contain tumor in well under \SI{1}{\percent} of draws, and the
network would rarely see a positive example. We therefore center
\SI{33}{\percent} of patches on a tumor voxel and a further \SI{33}{\percent}
on any labeled voxel, with the remainder sampled uniformly. Candidate centers
are indexed once per volume and cached.

\paragraph{Loss.}
A soft Dice term \cite{milletari2016v} over the foreground classes is summed with a class-weighted
cross-entropy (weights $0.2$, $1.0$, $4.0$ for background, liver and tumor, respectively).
Cross-entropy alone is dominated by background at this class imbalance.

\paragraph{Augmentation.} Random flips along all three axes and multiplicative
and additive intensity jitter.

\subsection{Inference and post-processing}

Whole volumes are predicted by sliding window over overlapping $128^3$ patches
with \SI{50}{\percent} overlap, blended with a Gaussian weight so that
patch borders, where the network has least surrounding context, contribute
least, avoiding seam artifacts.

Post-processing retains the largest connected component of the predicted liver
region, fills holes within it, constrains tumor predictions to lie inside it,
and removes tumor components smaller than $T=50$ voxels
(Section~\ref{sec:results-pp}).
Hole filling is required since hypodense necrotic lesions are frequently
assigned to background by the classifier, leaving a cavity in the predicted
liver exactly where the lesion lies.

\subsection{Evaluation}

We report per-case Dice: one score per volume, averaged over volumes, as is
conventional for LiTS. A volume containing no tumor in which no tumor is
predicted scores 1.0. Because this convention makes the reported mean sensitive
to binary outcomes on tumor-free volumes, we additionally report the mean
restricted to tumor-bearing volumes, and state both throughout.

Dice is computed without a smoothing term. With smoothing, a volume with no
ground-truth tumor and several thousand false-positive voxels scores a small
positive value rather than zero, which understates a categorical failure. We
additionally classify each volume as \emph{detected}, \emph{missed},
\emph{no overlap}, \emph{false positive} or \emph{empty correct}, which
distinguishes failure modes that a single averaged score mixes.

\FloatBarrier
\section{Results}

\subsection{Main result}

\begin{table}[htbp]
\centering
\caption{MSD Liver per-case Dice under five-fold cross-validation over all
\headlineNVol{} public volumes. Every volume is held out exactly once. No
checkpoint selection is performed: each fold trains for a fixed 200 epochs and
the final weights are scored, because the held-out volumes serve as both the
validation and the reported set. Post-processing applied as described in
Section~3.6. Intervals are 95\% percentile bootstrap over cases
(\num{20000} resamples), which is the uncertainty relevant to generalization;
see Section~\ref{sec:results-cv}. Parameter counts are exact.}
\label{tab:main}
\begin{tabular}{lrcc}
\toprule
Model & Parameters & Liver Dice & tumor Dice \\
\midrule
Ours (hybrid) & \num{536990} & \headlineLiver{} \headlineLiverCI{}
                             & \headlinetumor{} \headlinetumorCI{} \\
\bottomrule
\end{tabular}
\end{table}

\begin{table}[htbp]
\centering
\caption{Comparison on MSD Task03\_Liver. All rows are five-fold cross-validation
over the same \headlineNVol{} public volumes, with no separate held-out set, so
the figures are directly comparable. We omit nnU-Net's challenge-submission
entry, which was scored on withheld cases we have no access to. nnU-Net does not
report parameter counts for these configurations.}
\label{tab:benchmark}
\begin{tabular}{lrcc}
\toprule
Method & Params & Liver & tumor \\
\midrule
nnU-Net, 2D \cite{isensee2018nnu}                      & ---  & 0.9437 & 0.5394 \\
nnU-Net, 3D low-res only \cite{isensee2018nnu}         & ---  & 0.9469 & 0.4701 \\
nnU-Net, 3D cascade \cite{isensee2018nnu}              & ---  & 0.9538 & 0.5849 \\
nnU-Net, 3D \cite{isensee2018nnu}                      & ---  & 0.9411 & 0.6174 \\
nnU-Net, 3D + 3D cascade ensemble \cite{isensee2018nnu}& ---  & 0.9543 & 0.6182 \\
\midrule
Dense 3D U-Net (this work, single split) & \num{12946851} & 0.9140 & 0.4578 \\
\textbf{Ours (hybrid)}                   & \num{536990}   & \headlineLiver{} & \headlinetumor{} \\
\bottomrule
\end{tabular}
\end{table}

\subsection{Cross-validation and the size of the uncertainty}
\label{sec:results-cv}

\begin{table}[htbp]
\centering
\caption{Per-fold tumor Dice (all-volume LiTS convention). Fold means span
$0.487$ to $0.654$. A permutation test that reassigns the same
\headlineNVol{} per-case scores to folds at random reproduces this spread
exactly: the observed standard deviation of $0.0596$ sits on the null median of
$0.0597$ ($p=0.50$), and a fold as low as $0.487$ occurs in \SI{31}{\percent}
of random assignments. The variation between folds is therefore attributable to
which cases fell together, not to differences between the trained models.}
\label{tab:folds}
\begin{tabular}{lrrcc}
\toprule
Fold & Volumes & tumor-free & tumor (all) & tumor (bearing) \\
\midrule
0 & 27 & 3 & 0.5801 & 0.5276 \\
1 & 26 & 1 & 0.6543 & 0.6405 \\
2 & 26 & 3 & 0.5753 & 0.5634 \\
3 & 26 & 6 & 0.4872 & 0.4834 \\
4 & 26 & 0 & 0.5892 & 0.5892 \\
\midrule
Pooled & \headlineNVol{} & 13 & \headlinetumor{} & \headlinetumorBearing{} \\
\bottomrule
\end{tabular}
\end{table}

\begin{figure}[htbp]
\centering
\figfile[width=\textwidth]{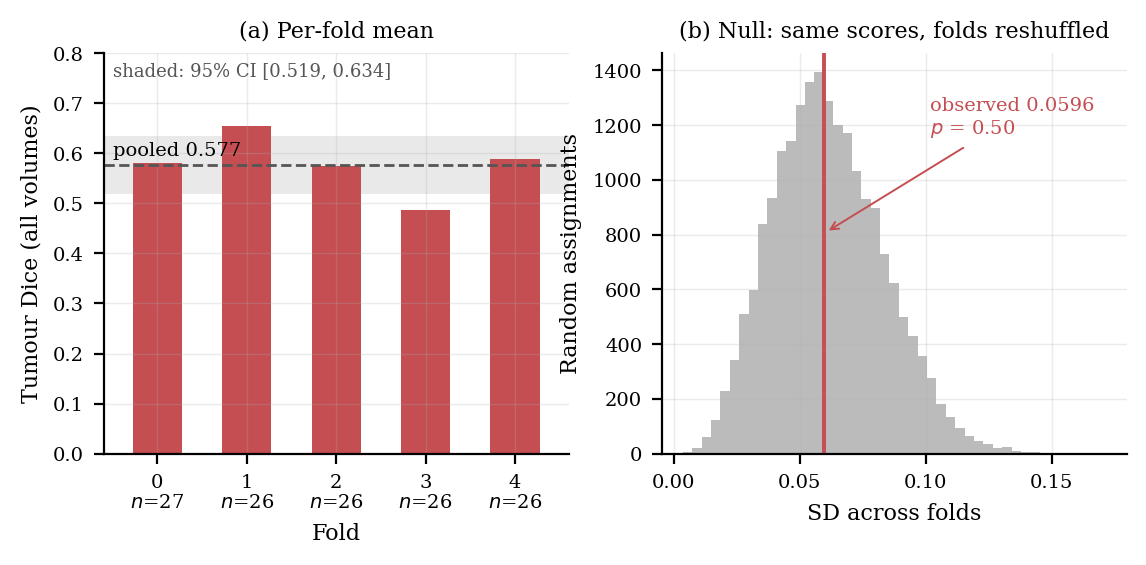}
\caption{(a) Per-fold tumor Dice, with the pooled mean (dashed) and its
\SI{95}{\percent} bootstrap interval over cases (shaded). (b) Permutation null:
the same \headlineNVol{} per-case scores reassigned to folds of the same sizes
at random, \num{20000} times. The observed standard deviation across folds falls
at the median of that null, so the fold-to-fold spread carries no information
about the trained models, it reflects only which cases fell together.}
\label{fig:folds}
\end{figure}

Table~\ref{tab:benchmark} places this against nnU-Net on the same task and the
same protocol. One difference should be kept in view. nnU-Net resamples to the dataset median
spacing, which for this task is $1.0 \times 0.77 \times 0.77$~\si{\milli\meter};
we resample to \SI{1.5}{\milli\meter} isotropic, giving roughly half the
in-plane resolution and \SI{5.7}{} times the voxel volume. Their
low-resolution stage is coarser still, a further factor of four on every axis,
or \SI{11}{} times our voxel volume. Against that low-resolution configuration our tumor Dice is higher
by $0.107$. Against their full-resolution 3D U-Net it is lower by $0.040$, a
difference contained within our confidence interval (\headlinetumorCI{}); our
liver Dice exceeds every nnU-Net configuration on this task. The model uses
\num{536990} parameters.

We report the volume-level mean, computed by pooling all \headlineNVol{}
per-case scores, rather than the mean of fold means. The two agree here to four
decimal places ($0.5773$ against $0.5772$), but they are not identical in
general because \headlineNVol{} does not divide evenly into five folds, and only
the pooled figure supports a bootstrap over cases.

The standard deviation
across folds ($0.0596$) describes how much a \num{26}-volume sample of this
dataset varies. The bootstrap interval (\headlinetumorCI{}) describes how much
the pooled estimate would move on a different sample of liver CT volumes.

Two further observations. First, the model produces a tumor Dice of exactly
zero on \num{16} of the \headlineNBearing{} tumor-bearing volumes, it detects
nothing at all in \SI{14}{\percent} of cases. Second, the
liver figure is stable across folds (case-level standard deviation $0.032$
against $0.334$ for tumor); essentially all of the uncertainty in the headline
is tumor uncertainty.

\subsection{Ablation: block variant}
\label{sec:results-3d}

\begin{table}[htbp]
\centering
\caption{Block variant ablation on MSD Liver: one seed per variant, identical
training, identical post-processing ($T=50$ voxels), evaluated on the same
held-out split of 19 volumes. Comparisons against the dense baseline are paired
over volumes. A single seed is used for every row so that the paired structure
is preserved; the three-seed figures for the hybrid variant are reported
separately in Section~\ref{sec:results-protocol}.}
\label{tab:ablation3d}
\begin{tabular}{lrccc}
\toprule
Variant & Parameters & Liver & tumor (all 19) & tumor (17 bearing) \\
\midrule
Dense 3D U-Net             & \num{12946851} & 0.914 & 0.458 & 0.512 \\
(a) Fixed shift            & \num{485987}   & 0.941 & 0.574 & 0.524 \\
(b) Learnable depthwise    & \num{557726}   & 0.945 & 0.596 & 0.548 \\
(c) Hybrid                 & \num{536990}   & 0.945 & 0.600 & 0.553 \\
\bottomrule
\end{tabular}
\end{table}

The clearest difference is on liver. The hybrid variant exceeds the dense
baseline by $+0.031$ (\SI{95}{\percent}~CI $[+0.013, +0.051]$; paired $t$
$p=0.006$, Wilcoxon $p=0.0005$ over all 19 volumes). On tumor, over the 17
tumor-bearing volumes, the difference is $+0.041$ (\SI{95}{\percent}~CI
$[+0.005, +0.081]$; paired $t$ $p=0.056$, Wilcoxon $p=0.064$). The bootstrap
interval excludes zero while neither test reaches $p<0.05$ at $n=17$. Among the
factorized variants, freezing \emph{every} spatial filter (variant a) is worse
than leaving them learnable by $-0.024$ ($p=0.025$), whereas freezing only the
first decoder stage (variant c) is indistinguishable from leaving all of them
learnable ($+0.005$, $p=0.744$).

Under the LiTS convention a volume with no tumor scores 1.0 when
nothing is predicted and 0.0 otherwise, so each of the two tumor-free volumes
is worth $1/19$ of the mean as a binary outcome. After identical
post-processing the dense model still emits \num{63} and \num{5394} spurious
tumor voxels on those two volumes and scores 0.0 on both; the hybrid model
emits none and scores 1.0 on both. The all-volume difference is therefore
predominantly a false-positive result, not a segmentation-quality one. This explains why the gap in all-volume tumor ($+0.142$) is four times the tumor-bearing gap.

That the \num{536990}-parameter model outperforms the \num{12946851}-parameter
dense network under identical conditions is, we suggest, a regularization
effect: with 93 training volumes, the dense model has capacity to overfit that
the factorized model does not. The false-positive pattern above is consistent
with this.

The post-processing threshold, $T=50$ voxels, was selected by
comparing test results using the factorized models
(Section~\ref{sec:results-pp}). It removes the hybrid model's small spurious
components and does not remove the dense model's \num{5394}-voxel one, which is
far above any plausible threshold.

\subsection{Where spatial capacity is used}
\label{sec:results-allocation}

\begin{table}[htbp]
\centering
\caption{Mean number of open taps per depthwise filter under $L_0$ gating
($\lambda=10^{-5}$), by network depth, on the 2D endoscopy benchmark. Three
seeds; per-layer standard deviation across seeds 0.16.}
\label{tab:taps}
\begin{tabular}{lccc}
\toprule
Seed & Encoder & Deep decoder & Shallow decoder \\
\midrule
0 & 5.47 & 1.94 & 3.80 \\
1 & 5.66 & 2.04 & 3.87 \\
2 & 5.47 & 2.17 & 3.90 \\
\bottomrule
\end{tabular}
\end{table}

\begin{figure}[htbp]
\centering
\figfile[width=\textwidth]{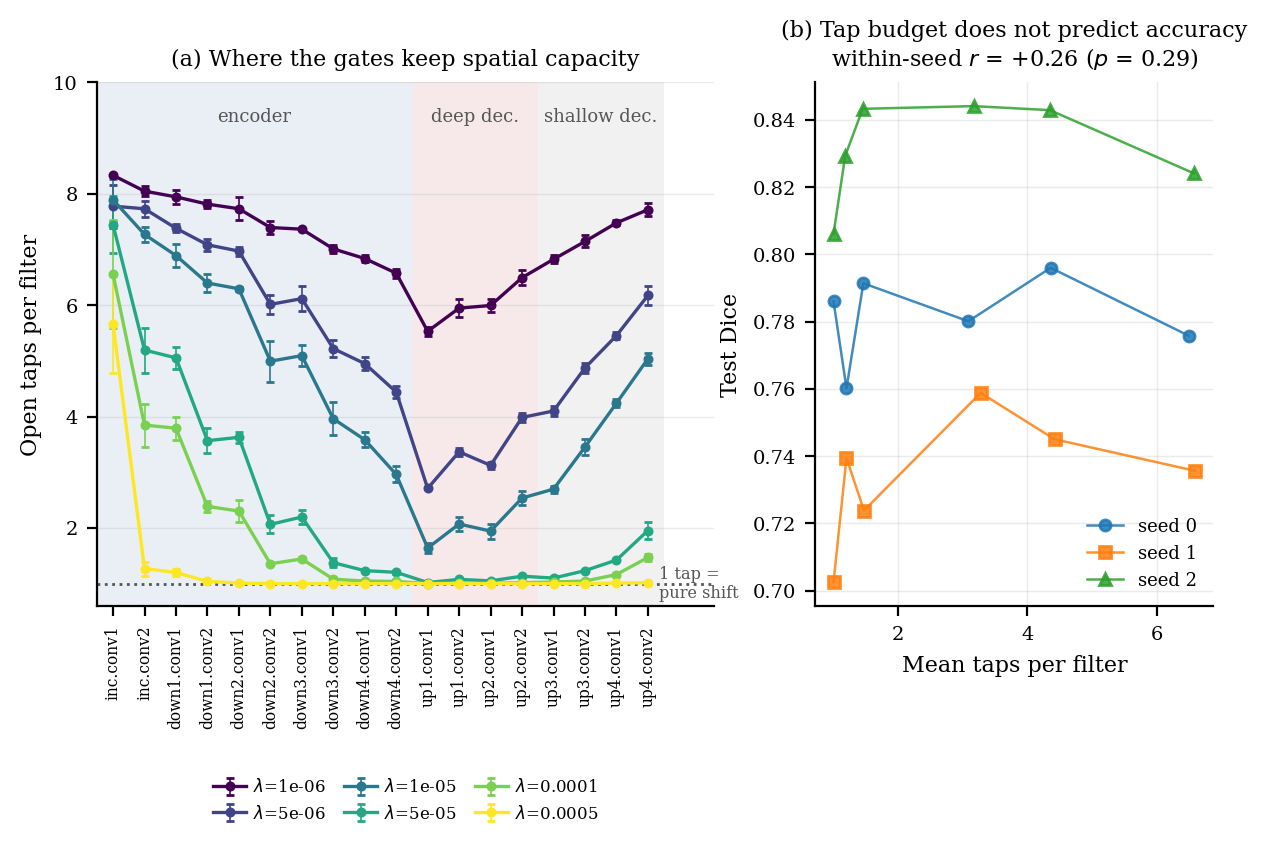}
\caption{$L_0$ gate allocation on the 2D endoscopy benchmark. (a) Mean open
taps per depthwise filter against network depth, at six penalty strengths, with
error bars over three seeds. The profile is V-shaped at every $\lambda$: the
gates retain spatial capacity in the shallow encoder, shed it fastest at
\texttt{up1.conv1}, the first decoder stage, fed directly by the
bottleneck, and restore it through the shallow decoder as resolution returns.
The ordering is preserved across a 500-fold range of $\lambda$ and across seeds
(mean per-layer standard deviation 0.16 at $\lambda=10^{-5}$, maximum 0.38).
Table~\ref{tab:taps} reports the grouped means at $\lambda=10^{-5}$.
(b) Total tap budget against accuracy, by seed. Reducing spatial parameters
sevenfold produces no detectable change, while the gap between seeds
($0.098$ Dice between seed means) is nearly twice the largest variation any
$\lambda$ produces within a seed ($0.056$).}
\label{fig:taps}
\end{figure}

The network sheds spatial capacity from the deep decoder first and retains it in
the encoder, reproducibly across seeds and consistently with increasing
$\lambda$ (Figure~\ref{fig:taps}a). The minimum falls on \texttt{up1.conv1} at
every penalty strength tested. This is the measurement that motivates variant~(c)
(Section~\ref{sec:whyup1}).

We observed that total tap budget, however, does not predict accuracy. Across the $\lambda$ sweep
the within-seed correlation between mean taps per filter and Dice is $r=+0.26$
($p=0.29$, $n=18$); reducing
retained spatial parameters from \SI{72}{\percent} to \SI{11}{\percent} produces
no measurable change. A separate sweep over which blocks to freeze, five
configurations at five seeds, likewise finds no differences
(one-way ANOVA $p=0.66$), while every partially frozen configuration exceeds the
fully frozen one.

We therefore state the finding as: \emph{some} learnable spatial filtering is
required, but a large majority of it is not, and the network's own allocation
does not identify a uniquely best placement.

\subsection{2D benchmark}
\label{sec:results-2d}

\begin{table}[htbp]
\centering
\caption{2D endoscopy benchmark (Kvasir-SEG \cite{jha2019kvasir}), Dice on a fixed held-out split,
mean $\pm$ standard deviation over three random initializations. The split is
held fixed across all runs (\texttt{split\_seed}~$=0$, independent of the run
seed) so that variation reflects initialization only, and comparisons against
the dense baseline are paired by seed.}
\label{tab:kvasir}
\begin{tabular}{lrc}
\toprule
Variant & Parameters & Dice \\
\midrule
Dense U-Net                & \num{17267393} & 0.805 $\pm$ 0.010 \\
Fixed shift, with padding projection & \num{4658780} & 0.798 $\pm$ 0.023 \\
Fixed shift, no projection & \num{1930433} & 0.794 $\pm$ 0.003 \\
Learnable depthwise        & \num{1978268} & 0.816 $\pm$ 0.022 \\
Hybrid                     & \num{1954076} & 0.811 $\pm$ 0.004 \\
\bottomrule
\end{tabular}
\end{table}

\begin{figure}[htbp]
\centering
\figfile[width=0.85\textwidth]{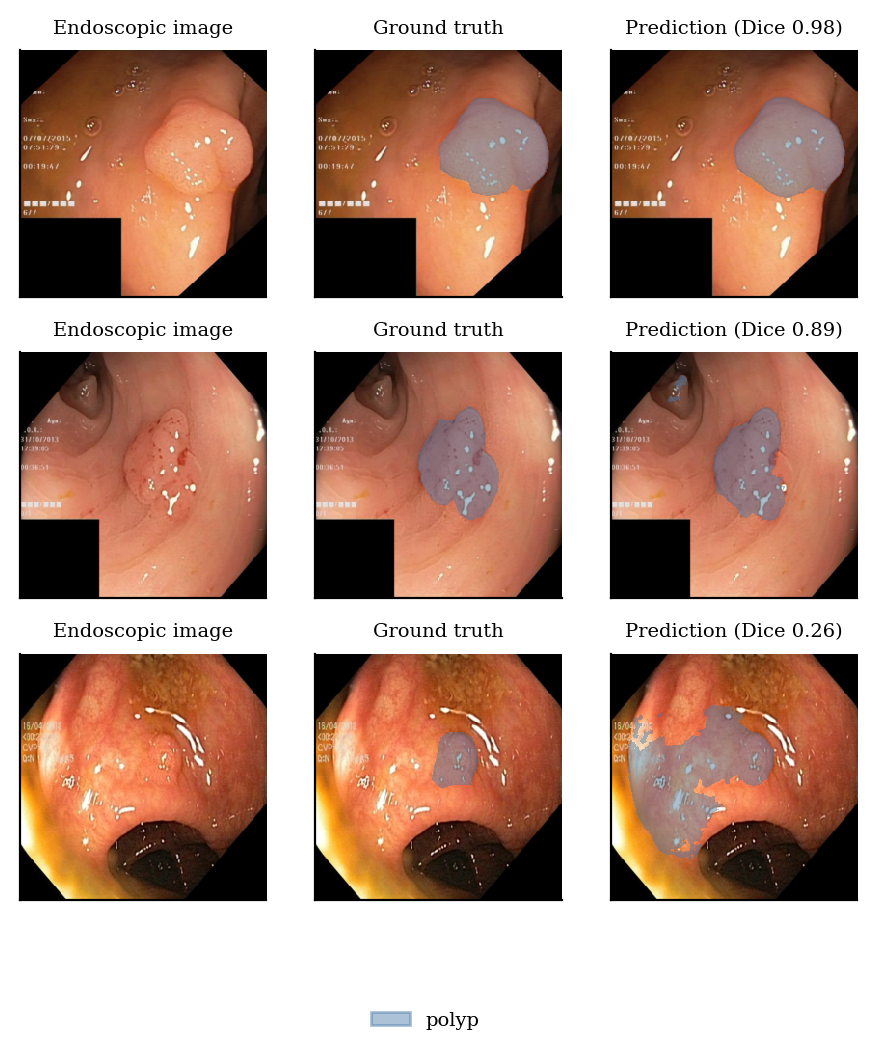}
\caption{Kvasir-SEG segmentations from the hybrid configuration
(\texttt{up1,up2,up3} frozen, \num{1954076} parameters), on the fixed test split
of 100 images (\texttt{split\_seed}~$=0$). Rows are the 5th, 50th and 95th
percentile of test Dice, ranked, so they are representative rather than
selected: $0.98$, $0.89$ and $0.26$. Across the split the mean is $0.800$ but
the median is $0.889$, with a minimum of $0.000$ and a maximum of $0.986$, as
on MSD Liver, most cases are segmented well and a tail of failures accounts for
most of the shortfall in the mean.}
\label{fig:kvasir}
\end{figure}

Three observations transfer from this benchmark to the 3D setting. First, the
channel-padding projection used in a common shift implementation accounts for
\SI{59}{\percent} of that model's parameters and produces no measurable accuracy
benefit: with the projection, \num{4658780} parameters and $0.798$; without it,
\num{1930433} parameters and $0.794$ ($p=0.81$). Second, the hybrid
configuration exceeds the fully frozen one by $+0.018$ ($p=0.014$) while
remaining indistinguishable from the fully learnable one, the same pattern
observed in 3D. Figure~\ref{fig:kvasir} shows representative segmentations.
Third, no factorized variant differs significantly from the
dense baseline under a paired test over the three seeds: hybrid $+0.006$
($p=0.32$), learnable depthwise $+0.011$ ($p=0.36$), fixed shift without
projection $-0.011$ ($p=0.11$). The hybrid configuration therefore matches a
dense U-Net using $8.8$ times fewer parameters. We note the contrast with the 3D
setting, where the factorized model exceeded the dense baseline
(Section~\ref{sec:results-3d}); with 800 training images rather than 93
volumes, the regularization advantage we attribute to factorization there is
correspondingly weaker here.

\begin{table}[htbp]
\centering
\caption{Freezing configuration sweep on the 2D endoscopy benchmark, five seeds
each, fixed split. Configurations are cumulative, beginning at the deepest
decoder stage. No configuration differs significantly from any other (one-way
ANOVA $F=0.61$, $p=0.66$), and all exceed the fully frozen model
($0.794 \pm 0.003$).}
\label{tab:cutpoints}
\begin{tabular}{lrcc}
\toprule
Frozen stages & Parameters & Filters frozen & Dice \\
\midrule
\texttt{up1}                        & \num{1964444} & \SI{29}{\percent} & 0.819 $\pm$ 0.011 \\
\texttt{up1,up2}                    & \num{1957532} & \SI{43}{\percent} & 0.810 $\pm$ 0.013 \\
\texttt{up1,up2,up3}                & \num{1954076} & \SI{51}{\percent} & 0.815 $\pm$ 0.008 \\
\texttt{down4,up1,up2,up3}          & \num{1944860} & \SI{70}{\percent} & 0.808 $\pm$ 0.015 \\
\texttt{down3,down4,up1,up2,up3}    & \num{1937948} & \SI{84}{\percent} & 0.814 $\pm$ 0.012 \\
\bottomrule
\end{tabular}
\end{table}

\subsection{Post-processing}
\label{sec:results-pp}

The size threshold $T$ was fixed at 50 voxels before the cross-validation was
run and held constant across all five folds and all model variants. It was selected by comparing results across five
settings on the held-out split of the earlier single-split protocol, not on a
separate validation set. We kept it in the cross-validation
scores, because the value affects the comparisons in
Table~\ref{tab:ablation3d} identically for every variant.

The effect of post-processing is concentrated rather than uniform. Across the
three initializations of the hybrid variant on the single split the mean gain
was $+0.036$ tumor Dice, made up of $+0.105$, $-0.000$ and $+0.004$: it removes
false positives when they are present and does nothing when they are not. The
same pattern separates the model variants in Section~\ref{sec:results-3d}, at
$T=50$ the hybrid model's spurious components are removed entirely while the
dense model's \num{5394}-voxel component survives.

\subsection{Failure analysis}
\label{sec:results-failure}

\begin{figure}[htbp]
\centering
\figfile[width=\textwidth]{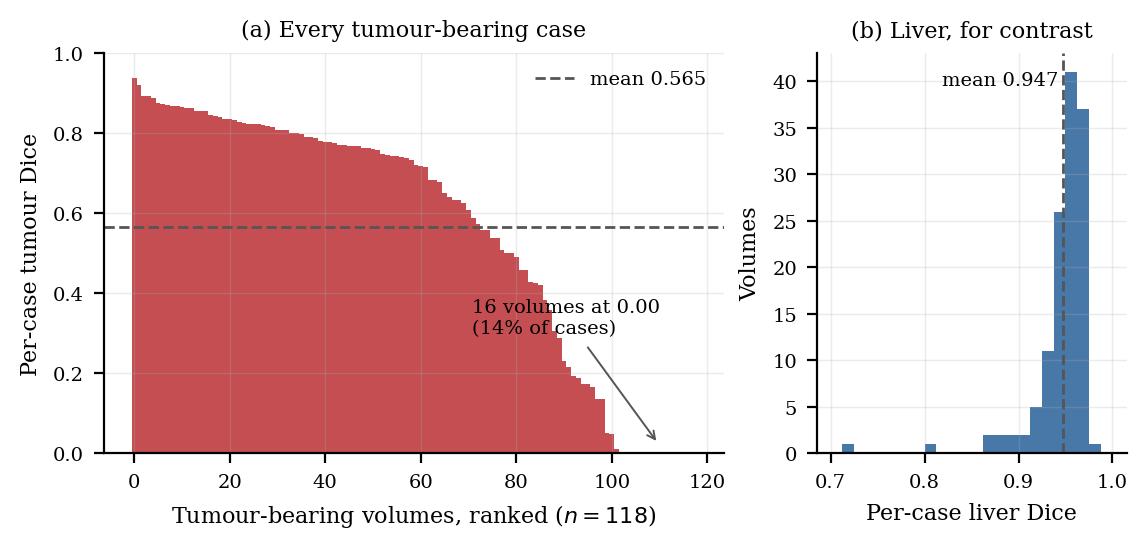}
\caption{Per-case Dice under five-fold cross-validation over all
\headlineNVol{} volumes. (a) tumor Dice for every tumor-bearing volume,
ranked. The error is not distributed around the mean: a body of adequate
segmentations is followed by \num{16} volumes (\SI{14}{\percent}) in which the
model detects no tumor at all, shown in grey. (b) Liver Dice for the same
volumes, on the same evaluation, for contrast, case-level standard deviation
$0.032$ against $0.334$ for tumor.}
\label{fig:percase}
\end{figure}

\begin{figure}[htbp]
\centering
\figfile[width=0.62\textwidth]{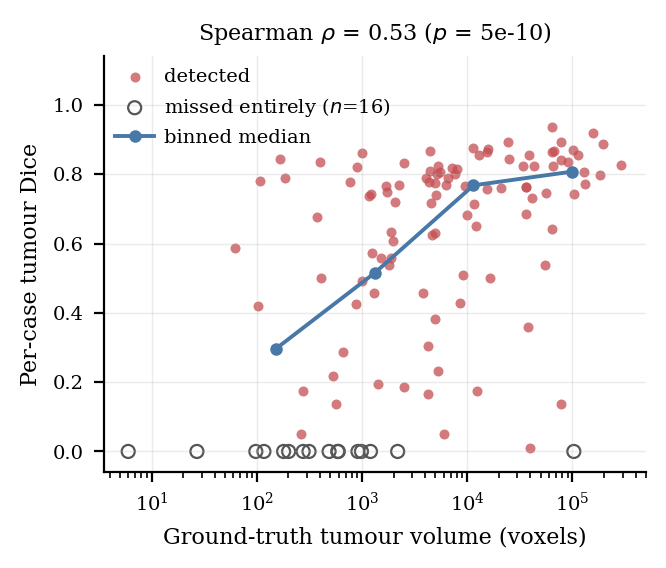}
\caption{Per-case tumor Dice against ground-truth lesion volume, log scale.
Open circles are the volumes missed entirely. Performance degrades
systematically with decreasing burden (Spearman $\rho=0.53$,
$p=5\times10^{-10}$, $n=\headlineNBearing{}$): the failures are concentrated in
small-lesion cases rather than distributed at random.}
\label{fig:burden}
\end{figure}

\begin{figure}[htbp]
\centering
\figfile[width=0.85\textwidth]{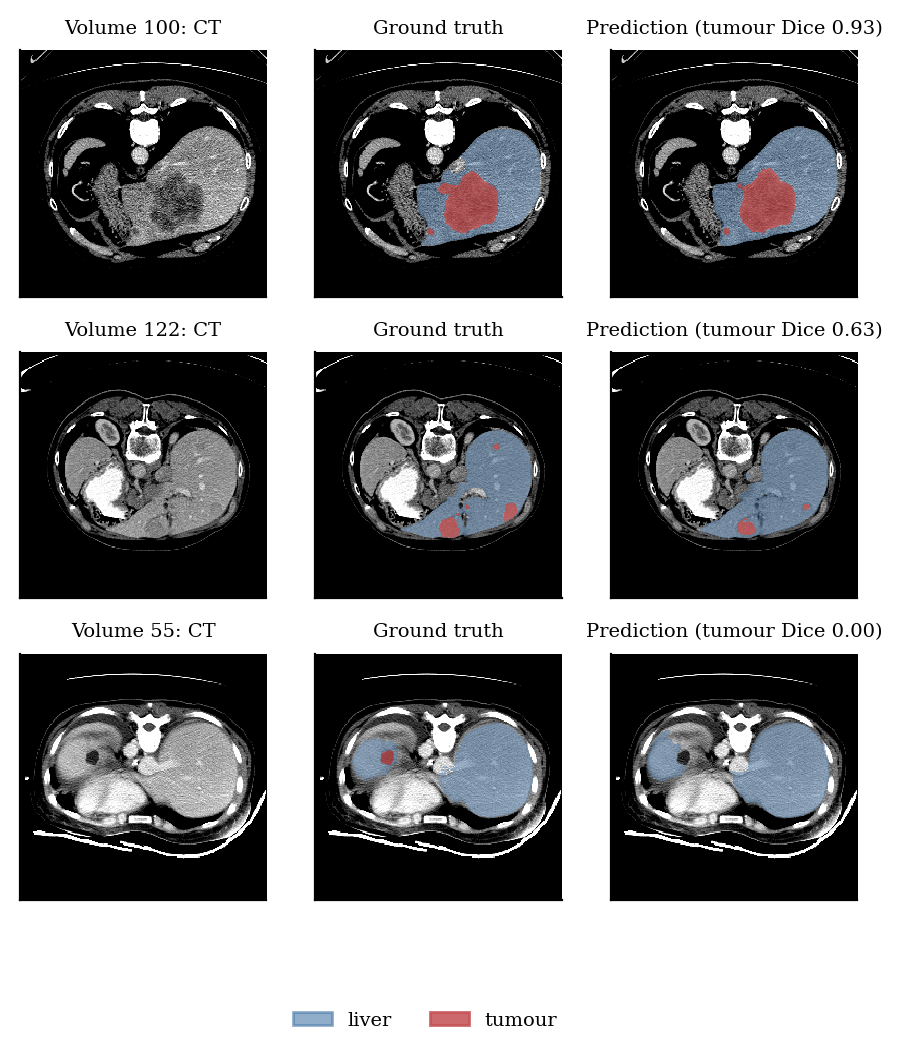}
\caption{Representative segmentations spanning the range in
Figure~\ref{fig:percase}. Each row shows one axial slice at the largest tumor
cross-section: CT left, ground truth center, prediction right. Top, volume 100,
a large confluent lesion, tumor Dice $0.93$. Middle, volume 122, multiple
smaller lesions of which one is recovered, $0.63$. Bottom, volume 55, a
\num{916}-voxel lesion for which the model predicts no tumor at all, $0.00$;
the liver is nonetheless segmented accurately at $0.95$. All panels use
a single fixed intensity window spanning the full preprocessed range,
$[-17, 201]$~HU, so brightness is comparable across the three volumes.}
\label{fig:qualitative}
\end{figure}

Per-case Dice on tumor is not uniformly distributed, and the failures have
identifiable and distinct causes. We characterize the five weakest test volumes.

\begin{table}[htbp]
\centering
\caption{The weakest test volumes, with contrast-to-noise ratio (CNR) between
tumor and surrounding parenchyma, and the fraction of ground-truth tumor
covered by the predicted liver region.}
\label{tab:failures}
\begin{tabular}{lrrrrl}
\toprule
Volume & CNR & Liver coverage & Recall & Precision & Interpretation \\
\midrule
39  & 0.02 & \SI{61}{\percent} & 0.00 & 0.05 & lesion isodense with parenchyma \\
104 & 3.65 & \SI{1}{\percent}  & 0.00 & 0.19 & hypodense lesion excluded from liver \\
55  & 5.17 & \SI{4}{\percent}  & 0.00 & 0.00 & hypodense lesion excluded from liver \\
49  & 1.07 & \SI{100}{\percent} & 0.69 & 0.11 & detected, over-segmented \\
21  & 1.07 & \SI{100}{\percent} & 0.42 & 0.42 & genuinely difficult \\
\bottomrule
\end{tabular}
\end{table}

Volume 39 contains a \num{103305}-voxel lesion at \SI{121}{HU} within
parenchyma at \SI{122}{HU}: a contrast-to-noise ratio of $0.02$. The lesion is
not distinguishable by intensity after windowing, and no architectural change
would recover it.

Volumes 104 and 55 present a different and instructive failure. Contrast is
high (CNR 3.65 and 5.17) and the lesions are large, but the predicted
\emph{liver} covers only \SI{1}{\percent} and \SI{4}{\percent} of the true
tumor. Under a mutually exclusive three-class softmax the network resolves a
large hypodense region inside the liver by assigning it to \emph{background};
the tumor head is then never given the opportunity to fire there. Hole filling
recovers the liver segmentation in these cases (volume 104: liver Dice
$0.823 \rightarrow 0.951$) but cannot relabel the voxels as tumor. A nested
region formulation, in which liver-or-tumor and tumor are predicted as
overlapping rather than competing classes, is a solution to this issue and is left to future work.

Volume 49 is the case post-processing is designed for: the lesion is found
(recall 0.69) but heavily over-segmented (precision 0.11).

Notably, tumor burden, which strongly predicted per-case Dice in our 2D
experiments on LiTS (Spearman $\rho = 0.83$), does not do so in 3D ($\rho = 0.09$).
Volumes with fewer than \num{2000} tumor voxels average $0.398$ and those above
\num{20000} average $0.483$. The move to 3D removes the small-lesion penalty
that dominates 2D slice-wise segmentation.

\subsection{Costs and benefits of parameter reduction}
\label{sec:cost}

\begin{table}[htbp]
\centering
\caption{Measured consequences of the parameter reduction. Inference timed on a
single CPU core at $128^3$, forward pass only; activation memory measured by
summing module output tensors during a forward pass at $128^3$, batch size 1,
base width 32. A ratio below one indicates the factorized model is worse on
that axis.}
\label{tab:cost}
\begin{tabular}{lrrr}
\toprule
 & Dense 3D U-Net & Ours & Ratio \\
\midrule
Parameters                      & \num{12946851} & \num{536990} & $24\times$ \\
Model file size (float32)       & \SI{49.4}{\mebi\byte} & \SI{2.05}{\mebi\byte} & $24\times$ \\
Training activation memory      & 4.58 GB & 6.06 GB & $0.76\times$ \\
CPU inference, s/patch          & 20.6 & 16.8 & $1.2\times$ \\
CPU inference, whole volume     & \SI{8.5}{\minute} & \SI{8.0}{\minute} & $1.06\times$ \\
\bottomrule
\end{tabular}
\end{table}

Activation tensors dominate memory
in 3D, and here the factorized model is the \emph{more} demanding of the two: a
depthwise separable block emits two intermediate tensors per layer, the
depthwise output and then the pointwise output, where a dense convolution emits
one. Fewer weights, more activations. And although the
factorized model performs far fewer multiply-accumulate operations, depthwise
convolutions have low arithmetic intensity, each weight is used once per
spatial position rather than reused across channels, so they are bound by
memory bandwidth and realize only a fraction of the theoretical speed-up. On a single CPU core a
$320 \times 239 \times 239$ volume takes \SI{8.0}{\minute} with the factorized
model against \SI{8.5}{\minute} with the dense one, a \num{24}-fold reduction
in parameters buying a \SI{6}{\percent} reduction in latency. Sliding-window
aggregation, which is identical for both models, accounts for part of the
remainder. On a typical multi-core workstation both fall to a few minutes.
Anyone selecting a model for inference speed should measure it rather than infer
it from a parameter count.

\subsection{Evaluation protocol matters more than is generally reported}
\label{sec:results-protocol}

On a 2D benchmark we observed a difference between two architectures
of $+0.050$ Dice, significant at $p=0.012$ over three seeds, which vanished
entirely ($p=0.49$) when the train/test split was held fixed and only the
initialization was varied. Seed variance in that setting was dominated by which
images fell in the test set rather than by the model.

\FloatBarrier
\section{Discussion}

\subsection{Deployment in hospital settings}

The practical utility for our proposed small parameter model is about distribution
and maintenance rather than raw speed. A model of approximately two megabytes
can be versioned in a repository alongside code, shipped to many sites without
special infrastructure, embedded in scanner or workstation software, and
retained for the multi-year periods that regulatory record-keeping requires. A
fifty-megabyte model can also be handled, but the difference compounds when
models are retrained per site, per scanner or per protocol, and when each
version must be archived.

Inference does not require a GPU. A single CPU core segments a volume in
approximately eight minutes and a typical workstation in a few minutes, within
memory available on ordinary hardware. This matters where GPU is lacking, and it makes retrospective batch analysis feasible on existing infrastructure.

\subsection{Why the factorization helps rather than merely costing less}

 With 93 training volumes, a
\num{12946851}-parameter 3D network has substantial capacity to overfit
training-specific structure. The depthwise factorization constrains each spatial
filter to act within a single channel, which is a strong structural prior, and
in this regime the constraint appears to help. Medical imaging datasets of one
to two hundred annotated volumes are typical rather than exceptional, so this
regime is the common case.

\subsection{Limitations}
\label{sec:limitations}

\paragraph{One initialization per fold.} The cross-validation trains one model
per fold at a single seed, so the headline figure carries no estimate of
initialization variance. On the earlier single-split protocol, three
initializations gave a tumor Dice of \splittumor{}; averaged over
\headlineNVol{} volumes rather than 19 that spread would be substantially
smaller, but we have not measured it.

\paragraph{Cross-validation covers one configuration.} The five-fold protocol
was run for the hybrid variant only. The comparisons between block variants in
Table~\ref{tab:ablation3d}, and against the dense baseline, remain paired tests
on a single split at one seed. Pairing over the same volumes controls for the
case-difficulty variation that dominates the fold spread, so those tests are
more sensitive than the interval in Table~\ref{tab:main} would suggest, but they
rest on \num{17} tumor-bearing volumes and should be read as such. We report an
absolute figure established under cross-validation and relative figures
established on a single split.

\paragraph{Resolution.} We resample to \SI{1.5}{\milli\meter} isotropic, coarser
than the $1.0 \times 0.77 \times 0.77$~mm used by nnU-Net's full-resolution
configuration, for reasons of storage. Our figures are therefore most fairly
compared against nnU-Net's low-resolution configuration.

\paragraph{Class formulation.} The mutually exclusive three-class softmax is
responsible for an identifiable failure mode: large hypodense lesions assigned
to background, excluding them from the predicted liver and preventing tumor
detection. Two of our nineteen test volumes fail this way.

\paragraph{Scope of the allocation finding.} We show that most learnable spatial
filtering can be removed without loss, and that removing all of it cannot. We do
not show that any particular allocation is optimal; a sweep over five freezing
configurations found no differences between them. All five configurations begin
at the deep decoder, following the gated allocation, so allocations that freeze
only shallow decoder stages remain untested.

\subsection{Future work}

The nested region formulation described in Section~4.6 addresses the clearest
identified failure mode and requires only a change of loss and output
parameterization. A cascaded arrangement, localising the liver, then
segmenting lesions within a cropped, effectively higher-resolution
volume, addresses the resolution limitation directly and is standard practice
on LiTS. Finally, the scaling argument of Section~\ref{sec:scaling} predicts
that the advantage of depthwise factorization grows with spatial dimensionality;
testing this on 4D data such as dynamic contrast-enhanced sequences is a natural extension.

\FloatBarrier
\section{Conclusion}

We show that depthwise separable factorization has a structural advantage in
three dimensions that it does not have in two: because the cubic kernel term
applies only to the depthwise stage, the same architecture grows by
\SI{5}{\percent} from 2D to 3D where dense convolution grows by
\SI{200}{\percent}. Building on this, we present a 3D U-Net of \num{536990}
parameters that reaches \headlinetumor{} tumor Dice and \headlineLiver{} liver
Dice on LiTS, comparable to published 3D configurations at approximately one
twenty-fifth of the parameters, and better than a dense 3D U-Net trained
identically on the same data. We further show that a large majority of the
network's learnable spatial filters can be replaced by parameter-free fixed
shifts without loss of accuracy, though not all of them, and we report the
measured rather than assumed consequences of parameter reduction: model size
falls by a factor of 24, but training memory rises by \SI{32}{\percent} and CPU
inference is only 1.2 times faster.

\section*{Data and code availability}
MSD Task03\_Liver is publicly available through the Medical Segmentation
Decathlon (\url{http://medicaldecathlon.com}) under CC BY-SA 4.0. This work
uses only the \headlineNVol{} publicly annotated training volumes; no data is
redistributed here.

Code, the fold assignment used for cross-validation, the per-case scores
underlying every reported figure, and the trained weights of the single-split
hybrid model are available at (\url{https://github.com/adham-synbio/3D_shift_UNet}).The fold assignment is deterministic and the reported per-case scores can be verified directly from the released CSV without retraining.

\bibliographystyle{plain}
\bibliography{references}

\end{document}